# Measuring Sentences Similarity: A Survey

**Mamdouh Farouk***

Department of Thermotechnik Computer Science, Assiut University, Markaz El-Fath, Assiut Governorate 71515, Egypt; mamfarouk@aun.edu.eg

## Abstract

**Objective/Methods**: This study is to review the approaches used for measuring sentences similarity. Measuring similarity between natural language sentences is a crucial task for many Natural Language Processing applications such as text classification, information retrieval, question answering, and plagiarism detection. This survey classifies approaches of calculating sentences similarity based on the adopted methodology into three categories. Word-to-word based, structure-based, and vector-based are the most widely used approaches to find sentences similarity. **Findings/Application**: Each approach measures relatedness between short texts based on a specific perspective. In addition, datasets that are mostly used as benchmarks for evaluating techniques in this field are introduced to provide a complete view on this issue. The approaches that combine more than one perspective give better results. Moreover, structure based similarity that measures similarity between sentences' structures needs more investigation.

## 1. Introduction

Recently, there is an explosion in the information on the Internet[1]. A massive amount of natural language data is added daily on the Internet. Moreover, the human literature in different cultures is digitalized and became available in digital libraries. A very large amount of this data formatted in natural language. This makes NLP techniques are crucial to make the use of this amount of data.

Similarity between sentences is represented by the degree of likelihood between these sentences. Moreover, finding sentences similarity is a crucial issue in much application. In other words, many natural language applications such as semantic search[2], summarization[3], question answering[4], document classification, and sentiment analysis, plagiarism[5] depend on sentences similarity. Furthermore, accuracy of measuring sentences similarity is a critical issue. Consequently, sentences similarity measuring has gained a lot of attentions. Many techniques have been proposed to measure similarity between sentences. This paper explores different sentences similarity approaches.

Sentences can be similar lexically or semantically. Lexical similarity means string based similarity. However, semantic similarity indicates similar meaning among sentence seven they have different words. From this definition the previously proposed approaches can be classified into string-based similarity and semantic similarity.

The string based or lexical based similarity approach considers the sentence as a sequence of character. Calculating similarity depends on measuring similarity between sequences of characters. Many techniques have been proposed in this class of similarity measure[6,7].

Semantic based similarity which depends on meaning of sentences has different approaches. These approaches adopt different techniques to compare two sentences semantically. The first approach, corpus based, finds the words similarity based on statistical analysis of big corpus. Moreover, deep learning can be used to analyze a large corpus to represent semantics of words[8]. The second approach, knowledge based, depends on handcrafted semantic net for words.

The meaning of words and relations between words has been included in this semantic net. The mostly used

**Author for correspondence*

semantic net is WordNet[9]. The third approach, structure based, uses structure information of a sentence to get the meaning of this sentence. Similar sentence should have similar basic structure. Differently, the new emerging approach, deep learning, is shown promising results in image processing field. This makes other fields to adopt deep learning technology. In Natural Language Processing, a learning model to learn word representation forms a very big corpus[8]. The generated vectors capture semantic features of words. Similar words should have closed vectors in the semantic space. Recently, many approaches for measuring sentences similarity exploit semantic word representation generated using deep learning.

This survey provides the reader with valuable information about the used methodologies in measuring relatedness between natural language short texts. In addition, it explains some popular techniques for measuring word-to-word similarity. Moreover, datasets that are used in evaluating sentences similarity techniques are introduced to enable the reader to build a complete background in this area.

There are many surveys that review sentence similarity issue[10-13]. Unlike other surveys, this survey distinguishes between words similarity methods and sentences similarity methods. Moreover, the presented techniques have been classified in taxonomy according to used methodology. The proposed taxonomy presents structure based text similarity. This type finds sentences similarity based on structure similarity of sentences. Moreover, using word embedding in calculating similarity is discussed in this survey.

The rest of this survey is organized as follow. Section two gives a brief overview on word-to-word similarity methods and their classification. Details of approaches used in measuring sentences similarity is explained in section three. Moreover, section four shows some public datasets that used in sentences similarity research area. Finally, the conclusion of the presented survey is reported in section five.

## 2. Words Similarity

Finding similarity between words is the core of sentence similarity. In the literature, there are many metrics for calculating word-to-word similarity. This section shows different approaches used to calculate similarity between words. A hierarchy for methods used for measuring words similarity is shown in Figure 1.

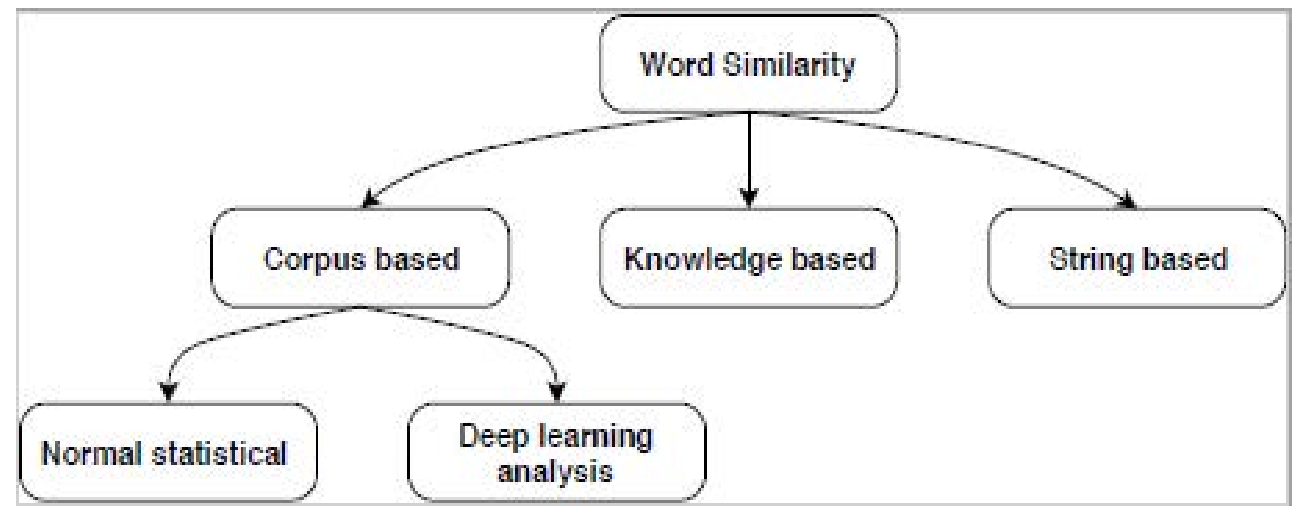

**Figure 1.** Different approaches for calculating word-to-word similarity.

There are three main approaches for measuring similarity between words. The first approach depends on analyzing a big corpus to find words similarity. The basic idea behind this approach is that semantic similar words appear in similar occurrence patterns. Moreover, two different techniques to analyze the corpus. The first uses normal statistical analysis while the other uses deep learning to find word semantic representation. The second main approach to find words similarity is knowledge based which depends normally on human crafted semantic network. This semantic network captures words' semantics and relations. Different ways to measure the words similarity based on semantic networks are proposed.

Furthermore, the third approach is terminological or string based which considers the word as a sequence of characters. It measures the similarity between these character sequences. The following subsections explain words similarity in details.

### 2.1 Corpus Based Words Similarity

A large number of the proposed approaches in words similarity are corpus based. In this type, valuable information is extracted from analyzing a big corpus. Moreover, analyzing words co-occurrence in a big corpus helps to assess similarity between these words accurately[14].

Furthermore, two different ways for statistical analysis of a corpus are exist. The first is using normal statistical analysis (such as LSA) and the second is using deep learning. In the first way, a big corpus is statistically analyzed through counting words in the corpus and documents. Calculating tf-idf, which is used as word weighting, is an important objective in the corpus analysis.

*Latent Semantic Analysis (LSA):* According to the assumption that co-occurrence words in the same context have similar meaning[15], many techniques proposed to measure words similarity. One of the most popular approaches in corpus based analysis is Latent semantic analysis (LSA)[12]. In LSA each word is represented by a vector based on statistical computations[16]. In order to construct these vectors a big text is analyzed and word matrix is constructed. In this matrix words and paragraphs are represented in rows and columns respectively. In addition, a mathematical technique called Singular Value Decomposition (SVD) is applied to reduce dimensionality. Based on the constructed word vector, the words similarity is calculated using cosine similarity.

*Using word embedding to find word similarity:* Deep learning is used to represent words semantically. A very large corpus is used for training to find word representation in a semantic space. The generated word representation depends on co-occurrence of words in the corpus. The idea of using deep learning is to train the model to guess a word given the surrounding words (Continuous Bag of Word model). Using this model a vector representation for words can be learned. There is another model (skip-gram) that tries to predict context words given a word. The generated vectors have size of 200 to 400 according to the parameters of the training process.

Moreover, using deep learning has shown good results in representing words semantically[8]. For example, if we make some calculations on the semantic vectors such as [Vect (*king*) – Vect (*man*) + Vect (*woman*)] the result will be very close to the vector represents the word *queen*. Cosine similarity between words' vectors is used to measure words similarity. Furthermore, word embedding is used widely to calculate words similarity[17].

*Web based word similarity:* This approach uses web content as a corpus. Similarity between words *W1* and W2, for example, is calculated as the ratio between the number of web pages that contains (w1 and w2) and number of web pages that contains only one word of them. For example, to calculate similarity between word p and word q, these queries "p and q", "p", and "q" are run on a web search engine. The similarity between words p and q is calculated according to the number of pages resulted from each query[18]. Moreover, snippets in search engine results can be used to improve similarity measurement. This is because snippets represent local context of both words[19].

## 2.2 Knowledge Based Words Similarity

In order to measure words similarity in this approach, an external resource such as WordNet[9] is used. In WordNet, English words are grouped into synsets. Each synset (group of synonyms) has a single meaning. These synsets are organized into a hierarchy constructing a semantic network in which semantic relations between synsets can be extracted easily.

Many approaches of finding words similarity use WordNet to measure relatedness between words. There are different methods to find word similarity using WordNet[20-23]. Sometimes these methods are combined calculate sentences similarity[24].

## 2.3 String Based Words Similarity

String based or lexical based word similarity depends on comparison between two sequences of characters. There are different methods to measure similarity between words based on string matching such as levenshtein distance, q-gram, and jaccard distance[25,26]. The following subsections give brief explanation for some of these methods.

*Levenshtein distance[27]:* distance between two strings is the minimal number of basic operations (insert, delete, or replace) needed to convert one string to another. For example, two operations are needed to convert "abcde" to "acrde" (delete and insert). The ratio between the distance and length of longer string is considered as the similarity between these strings.

*q-gram distance:* Q-gram distance[28] estimates the strings similarity based on how many common substrings of length q in both strings. For example, the distance between "abcde" and "acdeb" when q=2 is calculated as the sum of absolute differences between n-gram vectors of both strings.

## 2.4 Combined Methods

In[29] combine two approaches to measure similarity: corpus based and knowledge based.Simple average between the two measures is adopted. They proved that using combined measures of word-to-word similarity gives better results in the task of sentences similarity.

Differently[30] proposed another approach for measuring word similarity based on combining corpus analysis and background knowledge. Before analyzing corpus statistically, they make word sense disambiguation for corpus words. So every word in the corpus is attached to WordNet synset. In addition, to measure similarity between words shortest path distance in WordNet is used.

Moreover, the depth of words in WordNet hierarchical is used to represent level of specificity of words.

## 3. Sentence Similarity Methods

Different approaches for measuring similarity between natural language texts have been proposed. As shown in Figure 2, this survey classifies the methods used in similarity measuring according to the used technique into three main classes. The first category is word based similarity which considers the sentence as a set of words. It depends only on word similarity to calculate sentences similarity.

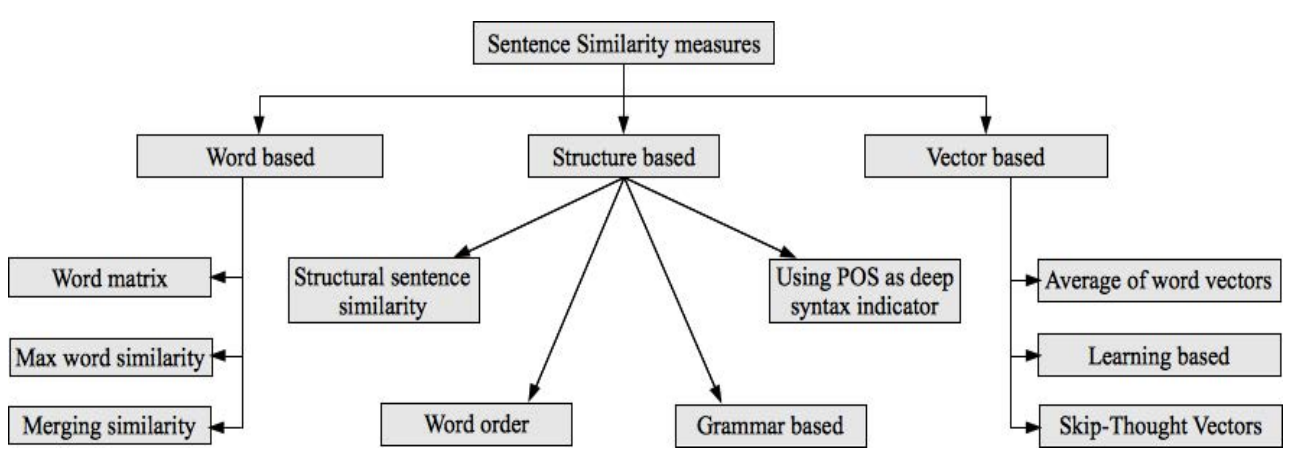

**Figure 2.** Semantic approaches for calculating sentences similarity.

The second class in our taxonomy is structure based similarity which utilizes structure information of the sentences in measuring similarity. Three different approaches to exploit syntactic information of a sentence: using grammar of the sentences, using POS, and using the order of words.

Vector based similarity is the third category in the proposed taxonomy. This kind of approaches is measuring similarity depending on calculating similarity between sentences' vectors. Details of each category of these approaches are mentioned in the following subsections.

### 3.1 Word Based Sentence Similarity

To measure relatedness between two sentences, this approach depends on word-to-word similarity. Sometime more than one method of word similarity is combined to measure similarity between sentences. Regardless the method for calculating words similarity, there are different ways that depend on words similarity to calculate sentences similarity.

#### *3.1.1 Max Similarity*

Sentences similarity in this method depends on coverage of a word in a sentence. Relatedness between a word and a sentence are calculated by measuring the similarity between this word and sentence words and the max similarity is selected according to the following equation.

Similarity between sentence S1 and sentence S2 calculated by the following equation:

In this equation, similarity between every word in S1 and the sentence S2 is measured and the average similarity is calculated[29].

#### *3.1.2 Similarity Matrix*

Another way to measure sentences similarity is to construct word similarity matrix through calculating the similarity between every pair of words in both sentences[24]. Using this similarity matrix and sentence binary vector (with element =1 if the word exist and 0 otherwise) the sentences similarity is calculated.

#### *3.1.3 Using Similar and Dissimilar Parts*

In[31] a new approach that uses similar parts of sentences and dissimilar parts to calculate. They use word embedding vectors to represent words. Using cosine similarity a words similarity matrix is constructed. Moreover, the authors construct semantic matching vector for each word using semantic matching function. There are different matching functions were proposed. The first considers the max match between word (*Wi)* in first sentence and every word in the second sentence. The second match function gets the weighted average between the word *Wi* and the words in the window in which the max match word in the center. The third match function considers the weighted average between *Wi* and all words in the other sentence. They make decompose to the generated match vectors to determine the similar parts and dissimilar parts for each vector. Based on the generated vectors similar matrix and dissimilar matrix are constructed. Finally, sentences similarity is computed after composing the similar matrix and dissimilar matrix.

#### *3.1.4 Using Word Sense Disambiguation*

Another proposed approach uses Word Sense Disambiguation (WSD) and synonym expansion[32] to measure sentences similarity. Each word is attached with a WordNet sense as a preprocessing step. A unit vector which contains all words included in both sentences is generated. Moreover, the original set of words of each sentence is expanded using WordNet synonyms. For each sentence a vector representation is constructed. The elements of this vector are calculated based on similarity between words in the unit vector and the set of expanded words of that sentence. Finally, a sentence semantic is calculated using cosine similarity of the two vectors.

### 3.1.5 Combined Similarity

Combine more than one word similarity measure gives better results. WordNet and word embedding similarities is combined to calculate the wordssimilarity[33].

Although the word based sentences similarity approach is clear and easy to implement, it ignores some important information such as structure of sentences.

Moreover, using structure information improves similarity measure. For example, attaching POS information to words guides the similarity measuring process.

This approach which considers a sentence as a bag of words is suitable for applications where sentences are poorly structured like tweets and social media text.

## 3.2 Structure Based Sentence Similarity

Considering syntactic information of a sentence facilitates understanding sentence meaning. Moreover, In short text, usually people uses similar structures to represent similar meanings[34]. This section explores some techniques that use structural information of sentences to improve the accuracy of measuring sentences semantic similarity.

### 3.2.1 Grammar Based

Lee et al proposed an algorithm calculates sentences similarity based on syntactic and semantic information of the sentences[35]. In the first step, the algorithm extracts grammar links or relations between words of each sentence. After a reduction process to the extracted links, a grammar matrix is constructed in which the rows are grammar links of the small sentence and the columns are grammar links of the second sentence. In the constructed matrix each link is associated with the words that are linked by this link. Semantic similarity of words that belongs to same type of links is measures using WordNet ontology. The final similarity calculated based on max value in each row of the grammar matrix. The main idea of this approach is finding similarity between words that have same type of links in the both sentences.

### 3.2.2 Using POS

Many techniques adopt Part-Of-Speech (POS) as syntactic information to calculate similarity between sentences. In[36] proposed a weighting strategy based on POS. They depend on the idea that certain POS and certain relation between POS are more important than others. In addition they combine the weighting strategy with bag-of-words approach to calculate semantic sentences similarity. Moreover, weights of different POS and weights of POS relation have been proposed in their work.

### 3.2.3 Using Word Order

Word order similarity is a way to assess sentences similarity considering order of words. Two sentences are typically similar if same words exist in both sentences in the same order. However, sentences should be considered as not completely similar if words of a sentence have different order as the other sentence. For example, consider these sentences: «the boy kills a woman» and «a woman kills the boy». These sentences have same set of words but they did not completely similar. Word order similarity is not used alone to measure sentences similarity. However, many approaches uses word order similarity combined with other methods[37].

An order vector which represents order of words in a sentence is constructed for each sentence. Using the constructed order vectors the word order similarity is computing[33].

The main advantage of structure based approach is exploiting structure information of sentences. People usually use similar structure to represent similar sentences. However, in many applications sentences are not well structured and not following grammar rules. In such cases this approach will not give the intended results.

## 3.3 Vector Based Sentences Similarity

In this approach the sentences similarity is assessed in two steps: 1. generation of vector representation for each sentence, and 2. measuring vectors similarity. Moreover, there are different approaches for representing a sentence by vector. These approaches try to capture features of sentences and represent these features in a vector. The two main categories for representing sentences by vectors are statistical based and learning based. Furthermore, the steps of calculating similarity between vectors also have different methods. The following subsections describe main approaches that use vectors to measure sentence similarity.

### 3.3.1 Distributional Sentence Similarity

The idea of this approach is inspirited from LSA. A matrix for counting sentence features is constructed. The rows of this matrix are corpus sentences and columns are features. The features may be unigrams, or may include bigrams or dependency pairs extracted through parsing the sentences. After constructing the matrix, TF-KLD is calculated as a weighting schema for feature. The purpose of using this weighting is to give higher weights to the more important features. This new weighting schema determines the weights of all features by calculating the probability of existence of the feature in paraphrased sentences and the existence of this feature in not-paraphrased sentences. Matrix factorization technique is used to extract sentences' vectors. Using these vectors sentence similarity can be measured[38].

### 3.3.2 Average of Word Vectors

In this method, a vector which represents a sentence is calculated as the average of words' vectors belong to this sentence. In[39] used the pre-trained GloVe word vectors. The corresponding vector, from GloVe semantic space, for each word in the sentence is grasped and the average between these vectors is calculated to represent the sentences. Cosine similarity between sentences' vectors is used to assess sentences semantic similarity.

### 3.3.3 Learning Based Vectors

Furthermore, depending on word vector representation generated using word embedding, Mueller in[40] proposed a training model to learn sentences similarity. A simple modification on Long Short Term Memory (LSTM) model is proposed and used in the training. In the input data, a sentence is represented bywords' vectors which have same size. Moreover, a label represents the human assessed similarity is assigned to each pair of sentences in the input data. The proposed Manhattan LSTM (MaLSTM) uses LSTM to read the input sentence vectors and uses the last hidden layer as a representation for the input sentences. The new generated representation for the sentences is used to learn semantic similarity measure.

### 3.3.4 Skip-Thought Vectors

This method gets vector representation for sentences based on unsupervised training model. The inputted sentences to this model are represented as a set of vectors. These vectors, which correspond to the words of the sentences, is selected from pre-trained word2vec space. The Skip-thought model is similar to Skip-gram model which predicts the surround words given a word. However, the Skip-thought model predicts the surround sentences based on the focused sentence. When calculating sentences similarity, first the sentences go for the pre-trained model to get vector representation. Then similarity is calculated based on a learning task.

Although vector based similarity approach has shown good results specially after using deep learning approach, capturing key features of sentences is still has many rooms of improvement.

Generally, many approaches combine between two or more of the explained techniques[41]. Table 1 shows some of the discussed approaches and the used techniques in each proposed system. In addition, the used dataset to evaluate each proposed system is mentioned with the achieved results. Table 1 shows the details of each approach and drawing the stat of the art. As shown in Table 1, the mentioned methods for measuring sentence similarity are used in a hybrid fashion.

Moreover, combining different approaches together to measure similarity between sentences has shown good results because it treats the problem from different perspectives. For example, combining word based approach and structure based produces a system that can work well for both well-structured sentences and not well structured sentences. Furthermore, using structural information in measuring similarity is not widely covered. Moreover, there are many rooms to investigate using sentence structure in calculating sentences similarity.

## 4. Similarity Datasets

This survey explains the most popular datasets in the area of sentence similarity. Moreover, because of the word similarity is used as basic step of similarity measure between sentences, a few datasets of word similarity measure are described as well. Moreover, beside the benchmark datasets there are many public available tools for sentence similarity such as[42,43].

### 4.1 Word Similarity Dataset

The following datasets used to evaluate words similarity techniques. Moreover, these datasets may be used to evaluate the semantic word representation approaches.

### 4.1.1 The WordSimilarity-353 Test Collection

This dataset contains two sets of data[44]. The first set contains 153 pairs of words and the other set contains 200

**Table 1.** Summary of sentence similarity approaches and the used datasets

| Method | WordNet | Word embed-ing | LSA | Word-order | Using struc-ture | Vector based | Trained-net-work | Data set | results |
|---|---|---|---|---|---|---|---|---|---|
| Atish 2018[30] | √ | | | √ | | √ | | Pilot | 0.837 |
| Distri-butional sentence similarity[38] | | | √ | | | √ | √ | MSRP | 80.41 |
| Grammar-based[35] | √ | | | | √ | | | MSRP Pilot | 71.02 -- |
| skip-thought | | √ | | | | √ | √ | MSRP SICK | 75.8 0.8655 |
| Wang[31] | | √ | | | | √ | √ | MSRP | 78.4 |
| Max similarity[29] | √ | | √ | | | | | MSRP | 70.3 |
| Similarity matrix[24] | √ | | | | | √ | | MSRP | 74.1 |
| WordNet and Word embedding | √ | √ | | √ | | | | MSRPPilot | 71.6 0.852 |

pairs of words. The combined dataset contains both sets (353pairs of words). Each pair of words assigned a human assessed similarity value between 0 and 10. 0 means not related words. While 10 means typical words. The human assessed values are assigned by many subjects. The data set is available in 'csv' format and tab delimited format. Many of the previously proposed approaches have used this dataset to make result evaluation[45,46].

### 4.1.2 Mturk-771 Test Collection

Mturk-771 dataset[47] contains 771 pairs of word with human judge relatedness score. It can be used for training and/or testing in case of supervised algorithms. The rating of relatedness is ranged from 1, means not related and 5 which means highly related. More than 20 ratings is collected for each pair in this dataset. Moreover, a refinement methodology has been applied to purify this dataset and eliminate the extreme values. Finally, the means for the ratings is assigned to each pair of words. This dataset is used in many approached to evaluated relatedness measure[48].

### 4.1.3 Rubenstein and Good Enough - RG-65

One of the earliest datasets in word similarity isRG-65 test collection[48]. It contains 65 word pairs annotated by human similarity measure. The labeled similarity measures are the means of assessments made by 51 subjects. The similarity attached to each pair is ranged from 0 to 4 (the higher the more similar).

## 4.2 Sentences Similarity Datasets

Generally, the available datasets that are used in evaluating text similarity measures are manually created. It contains pairs of sentences in natural language that are labeled by human similarity assessment. Normally, more than one human participate to assess the similarity. Furthermore, the created dataset that is used as academic benchmark has gained special care. For example, training of human raters, many raters for same sample, raters may annotate their rate by level of confidence, and agreement of multiple raters[50]. Moreover, a refinement methodology may adapt to produce the final dataset.

#### *4.2.1 Microsoft Research Paraphrase Corpus*

MSRP dataset[51] is published by Microsoft research center. This dataset contains around 5700 pairs of sentences (4000for training and about 1700 pairs for testing). Each pair of sentence is labeled by 0 (means dissimilar) or 1 (means similar). These sentences have been extracted from web news sources. The labels of pairs of sentences have been evaluated by human. This dataset is widely used in evaluating similarity measure techniques[32-38,42,43].

#### *4.2.2 Short Text Semantic Similarity Benchmark[52]*

This dataset is developed early to evaluate semantic similarity measure[53]. Originally it was created to measure words similarity and contained 65 pair of words. However, Li et al added definitions of each word using the Collins Co build dictionary to use this dataset in sentence similarity. Each pair of sentences is assessed by 32 English native speakers according to the similarity degree. This dataset has been adopted as de facto gold standard dataset[2]. Moreover, this dataset is used widely to evaluate similarity measure techniques[42]. Moreover, Pearson correlation[54] normally used to measure the relationship between human assessment and the proposed system results.

#### *4.2.3 SICK Dataset*

Sentences Involving Compositional Knowledge (SICK) dataset[55] is used in the shared task EemEval 2014. It contains 10000 pairs of sentences. Each pair is labeled by value between 1 and 5. This value represents the degree of relatedness between the sentences. This dataset is used as benchmark to evaluate technique of sentences similarity[40].

Moreover, there are many other datasets that used in this field such asQAsnt[14], WikiQA[56] which is new public available for question answering and sentences paring in open domain.

## 5. Conclusion

This study explores many approaches that are used to measure similarity between sentences. Approaches for measuring similarity between words are explained as a pre-request for sentences similarity task. Moreover, analyzing big corpus to find word co-occurrence is used to measure word-to-word similarity. Moreover, using deep learning approach to learn word embedding from a big corpus contributes to measuring word similarity. Knowledge based similarity approach uses external resource such as WordNet to measure word similarity. Differently, the presented approaches for measuring sentences similarity are classified into three main categories according to the used methodology. Word similarity based approach which depends on word-to-word similarity to calculate sentence similarity. Moreover, some approaches use sentence representation to measure similarity between sentences exploiting sentences structure shows improving in sentence similarity measuring. Combining between different approaches normally gives better results because it considers different aspects (lexical, syntactic, and semantic) of sentence. Moreover, the most widely used dataset for words similarity task and sentences similarity task are introduced.